# Metaphorical Language Change Is Self-Organized Criticality

Xuri TANG    Huifang YE

Huazhong University of Science and Technology

**Abstract**: One way to resolve the actuation problem of metaphorical language change is to provide a statistical profile of metaphorical constructions and generative rules with antecedent conditions. Based on arguments from the view of language as complex systems and the dynamic view of metaphor, this paper argues that metaphorical language change qualifies as a self-organized criticality state and the linguistic expressions of a metaphor can be profiled as a fractal with spatio-temporal correlations. Synchronously, these metaphorical expressions self-organize into a self-similar, scale-invariant fractal that follows a power-law distribution; temporally, long range inter-dependence constrains the self-organization process by the way of transformation rules that are intrinsic of a language system. This argument is verified in the paper with statistical analyses of twelve randomly selected Chinese verb metaphors in a large-scale diachronic corpus.

**Key words**: metaphorical language change; self-organized criticality; fractal

# 1 Introduction

Metaphor is an important mechanism of semantic change in a language (e.g. Burridge and Bergs 2017: 64; Croft 2000: 240; Sweetser 1990). For instance, the metaphor SOCIAL ORGANIZATIONS ARE PLANTS hypothetically motivates metaphorical expressions such as (1-4):

(1) *They had to **prune** the workforce.* (Croft and Cruse 2004: 205)

(2) *Employers **reaped** enormous benefits from cheap foreign labour.* (Croft and Cruse 2004: 205)

(3) *He works for the local **branch** of the bank.* (Croft and Cruse 2004: 205)

(4) *There is a **flourishing** black market in software there.* (Croft and Cruse 2004: 205)

But the actuation problem of language change (Weinreich et al. 1968: 102; Campbell 2008: 219) motived by metaphors — why the change occurs at the particular time and place that it does, how it proceeds, and what carries it along — requires further exploration. Gibbs (2021: 24) rephrases the problem as follows: what explains the emergence and meanings of these metaphorical expressions in the time and place in which they are employed?

The problem has been recently addressed by several studies largely within the framework of complexity theory and dynamic systems. The discourse dynamics approach to metaphor (Cameron and Deignan 2006; Cameron 2007; Cameron et al. 2009; Cameron and Maslen 2010; Gibbs and Cameron 2008) regards a metaphor as a temporary stability in language use. In continually-changing discourses, metaphorical expressions are inter-connected (such as the connection between *an invisible **enemy*** and *a **flaw in the system*** used in a discourse describing terrorism as *a sneaky way* [Cameron, 2010]), and they self-organize into systems across levels and timescales. The dynamic view of metaphor (Cameron 1999a; Gibbs and Cameron 2008; Gibbs 2010; Gibbs and Colston 2012; Gibbs and Santa Cruz 2012; Gibbs 2013, 2021; Müller 2008; Müller and Schmitt 2015; Tang 2021) advocates that metaphorical expressions are constrained in a complex, non-linear fashion by both cognitive/embodied and social/cultural forces operating on different time scales, and that metaphoric meaning is an emergent product of human self-organizing system. Based on the career-of-metaphor hypothesis (Bowdle and Gentner 2005), Yung (2021) uses a computational model to study the emergence of the THINGS ARE NETWORKS metaphor (as in *a network of hospitals*) and discovers that its linguistic forms vacillate between simile (*X is like a network*) and rhetorical metaphor (*X is a network*) in its early years, but it is primarily expressed as a categorization (*political networks*) since the beginning of 1900s.

Nevertheless, a successful resolution of the actuation problem should possess the power of prediction, specifying general laws and antecedent conditions, and predicting the course of subsequent events (Walkden 2017). Accordingly, the resolution of actuation problem of metaphorical language change requires not only descriptions of inter-connection and specification of constraints, but also general laws and rules with predictive power. Such laws can be a holistic and statistical profile of metaphorical expressions for each metaphor, which aggregates individual metaphorical expressions of a metaphor, depict its “supra-individual” (DiMaggio 1997) knowledge, and profile its presence in sociology (Yung 2021). Rules with antecedent conditions are what Gell-Mann (2002) calls schema that govern the emergence order of metaphorical expressions. These rules can be used to explain why Croft and Cruse (2004) arrange (1-4) in the order and predict people’s future metaphorical behavior such as the acquisition order of metaphorical constructions. Such knowledge can also clarify complexities confronting computational metaphor studies, such as metaphor identification that is difficult and theory-dependent (Ritchie 2013), personalized metaphor computation (Rai and Chakraverty 2020), and the construction of metaphor knowledge database (e.g. Rosen [2018], Stowe and Palmer [2018]).

This paper proposes such a holistic profile of metaphorical language change within the theoretical framework of Self-Organized Criticality (SOC), one of the most inspiring concepts in the development of complexity science (Watkins et al. 2016) and a general organizing principle governing dynamical systems with interacting degrees of freedom (Bak et al. 1987; Bak et al. 1988). On the basis of the view of language as complex systems (Beckner et al. 2009; Cameron and Larsen-Freeman 2007; Gromov and Migrina 2017; Massip-Bonet 2013; Paterson 2012) and the dynamic view of metaphor, this paper argues that metaphorical language change possesses the genotype, i.e. the sufficient ingredients that constitute a mechanism to produce a SOC state, and the phenotypes, the observable characteristics that are necessarily exhibited by SOC. In perceiving metaphorical language change as SOC, it is predicted that expressions of a metaphor should self-organize into a self-similar and scale-invariant fractal with a power-law distribution, and that they are constrained by long-range inter-dependence in the form of general and intrinsic transformation rules in a language system. This proposal is verified in the paper with empirical evidence collected from investigation of twelve randomly selected Chinese verb metaphors in a large-scale diachronic corpus.

## 2 Qualifying metaphorical language change as SOC

The genotype that enables a system to generate a SOC state consists of the following two ingredients (Watkins et al. 2016; Kafatos 2016):

– Non-linear interaction among composing entities, which is manifested as richness and diversity in the modes of interaction among the entities;

– Intermittent avalanche changes due to multiplicative interactions among neighboring entities and interaction threshold that only allows a response when some local dynamical variables exceed a value.

That metaphorical language change possesses the above ingredients can be naturally drawn from earlier studies based on the view of language as complex systems, the more recent view of language as self-organized criticality (Gromov and Migrina 2017) and the dynamic view of metaphor, which is detailed below.

## 2.1 Non-linear interaction in use of metaphor

Non-linear interaction among composing entities is the key-ingredient of SOC. In the view of language as a complex adaptive system, a language is a system consisting of multiple speakers (i.e. entities) interacting with each other, and that the past and present interactions among multiple speakers in the speech community feed forward into future behavior, motivating language change (Beckner et al., 2009). Because all linguistic sub-levels can be regarded as complex systems of their own (Geert and Verspoor 2015: 539) and metaphorical language is a sub-category of language use, interactions among speakers constitute the essential mechanism that motivates metaphorical language change.

Nonlinearity interaction in language is observed in language speakers' adaptive behavior in selecting from diversified linguistic patterns. One example in point is the bunch of English verbs describing horse movement: *walk*, *trot*, *canter*, and *gallop*, the usage of these verbs being determined by context, speakers, and time (Cameron and Larsen-Freeman, 2007). Multiple factors participate in the adaptive selection, including linguistic coherence (Nowak et al. 2002), social values such as prestige and social relation (Croft 2000: 32), and relevance (Van Orden et al. 2011). In addition, each speaker's exposure to linguistic experience is unique (Bybee 2006) and general social process generates orderly heterogeneity of language (Weinreich et al. 1968). Consequently, diversity is intrinsic in language (Beckner et al. 2009) and the production of an utterance involves an extremely complex recombination of elements from a great range of utterance patterns (Croft 2000: 29).

Accordingly, nonlinearity is observed in metaphor complexity, which is detailed in the dynamic view of metaphor. A metaphor can take diversified syntactic patterns. For example, the syntactic patterns of the metaphor SOCIAL ORGANIZATIONS ARE PLANTS surely are far more versatile than (1-4), each foregrounding one aspect of mapping between the source domain and the target domain, and each performing a different syntactic function. For instance, the verb *prune* in (1) foregrounds the mapping FORCEFUL REDUCTION IS CUTTING, while the verb *reap* in (2) foregrounds the mapping BENEFITING IS CUTTING. In (3) the vehicle *branch* functions as the object of a preposition, and in (4) the adjective *flourishing* is a noun modifier.

Metaphor complexity is also manifested in the observation that there is no overarching mechanism that decides the process of formulating or interpreting a metaphor (Gibbs and Colston 2012: 124). Metaphors are tremendously varied and should be described and explained with different theoretical and analytic frameworks (Cameron 1999b). For instance, the interaction theory (Richards 1965; Black 1955) might be suitable to explain anomalous collocation *prune the workforce* in (1), while the theory of conceptual metaphor well explains (3).

One more manifestation of metaphor complexity is the dimensions and factors in explaining a metaphor. Cameron (1999a: 8) notes that even within one analytic framework it may also need to be multi-dimensional to account for different aspects of metaphor. The multiple constraints in metaphorical interaction include evolutionary forces, cultural conditions, social context, knowledge of language, bodily states, and motivations etc. (Gibbs 2011; Gibbs 2013), or bodily experience, culture, cognition, and other things as are specified in Kövecses (2005).

## 2.2 Intermittent avalanche change

Metaphorical language change, like other types of language change, occurs in the form of intermittent avalanches, i.e. "at the fast time scale within an avalanche and the slow time scale between avalanches" (Watkins et al. 2016: 21).

A metaphor-motivated change proceeds at fast time scale once its development reaches a point, like an avalanche. It can be modeled with S-Curve (Bailey 1973; Blythe and Croft 2012; Tang et al. 2016). As illustrated in Figure 1, the change is almost invisible before Point A, but it becomes momentous after the point and proceeds at a fast rate, and finally it tails off slowly before reaching completion (Bailey 1973). The nature of avalanche in S-Curve is manifested by the phase between point A and B in the figure. Within this phase, the change is propagated very momentously in the speech community.

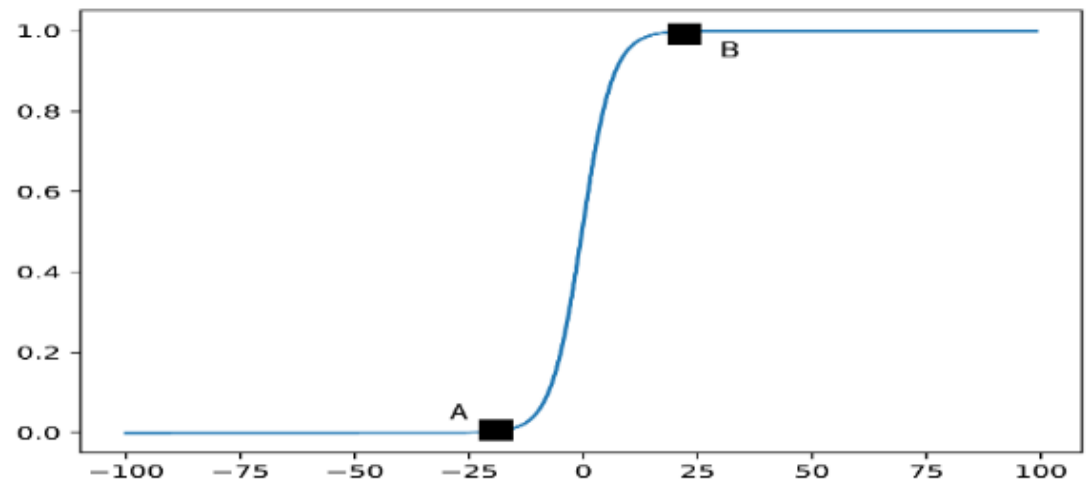

**Figure 1 The S-curve and avalanche in language change**

Avalanche metaphorical language change is resultant of multiplicative interaction in language systems, i.e. the interaction among multiple neighboring entities (language speakers) (Holden et al. 2009; Van Orden et al. 2003 ; Van Orden et al. 2011). In the language change field, multiplicative interaction can be conceptualized in the form of interaction in social networks discussed in Blythe and Croft (2012), Milroy (1980) and Milroy and Llamas (2013). Social networks are parts of structured, functional institutions such as classes, castes or occupational groups in which people interact meaningfully as individuals (Milroy 1980: 45-46). Speech events that occur among members of a social network are generally multiplicative, often involving more than two individuals.

By analogy with the general model in Jensen (1998), the multiplication of metaphorical language change can be described as follows. A novel metaphor is formulated by a random individual speaker in a social network. This event will influence the speaker him/herself and a group of people in his neighborhood, forming some sort of random sub-network. This sub-network will be modified, correlated with events of linguistic communication in the social network, but the sub-network should be below threshold. As the sub-network expands due to more use of the metaphor, more communicative events should spark off activity so that the threshold is overcome and the metaphor is propagated in the social network and is used in various linguistic structures and contexts, bursting out in the form of avalanche. Due to multiplication, nearest neighbor interactions extend to the periphery of the social network, or to other social networks, showing emergent behavior (Van Orden et al. 2003).

Metaphor-motivated changes are intermittent, meaning that the occurrences of these changes are not immediately sequent in time, but are separated at different time scales, and are of different stages. This can be exemplified by the tipping point metaphor introduced in van der Hel et. al. (2018) as the warning for abrupt and possibly irreversible changes in the climate system. They examine the text data of the metaphor between 2005 and 2014 and identify "four partly overlapping episodes, characterized by distinct linguistic and discursive uses of the metaphor across science and the news media" (van der Hel et al. 2018: 610): as a rhetorical device with deliberate metaphorical language among scientific peers in 2005-2007, as a metaphorical scientific concept by journalists in the same time, as a theory-constitutive metaphorical model for research from around 2007, and as conventional ideas and expressions for important impending change from around 2011. In regard to (1-4), Croft and Cruse (2004: 205) propose that these examples represent four stages in the development of the metaphor SOCIAL ORGANIZATIONS ARE PLANTS. Yung (2021) also identifies at least two stages of the THINGS ARE NETWORKS metaphor. Distinct stages in metaphorical language change are manifestations of its intermittence.

The nature of intermittence is consequent upon the threshold in language speakers' simulation of a metaphor. In language interaction, a speaker will adaptively decide whether to simulate a metaphorical expression, under the influence of the sociolinguistic and non-referential values (Croft 2016: 80; Grondelaers et al. 2010: 999) and the aim to maximize the relevance of the processed information (Sperber and Wilson, 1995). Because more frequent modes of expressions are the default choice unless there is a good reason for choosing something else (Halliday 2013), a novel metaphorical expression will not be selected unless it can achieve communicative purposes that other options cannot. This habitual tendency forms a threshold that upholds the use of metaphorical expressions. Take again the tipping point metaphor as the example. The threshold is observed in that the speaker (Professor Hans Joachim Schellnhuber) tried several metaphorical phrases (such as *switch-and-choke points* and *large-scale discontinuities*) in the interview but only the tipping point metaphor was picked up by the BBC journalist in 2005. It is also observed in that the tipping point metaphor is criticized before its acceptance for its tone of alarmism, the deterministic nature that human beings are defenseless against climate change and overly simplistic extension of the original tipping point concept (van der Hel et al., 2018). In the sandpile model (Bak et al. 1987; Bak et al. 1988), adding one grain of sand to the sandpile does not always trigger avalanches in the sandpile. Analogically, the use of one metaphorical expression by one individual may not lead to propagation of the metaphor. The threshold must be overcome, some local dynamic variables much exceed a value, and some preconditions must be met before a metaphorical expression is simulated.

To recapitulate, as natural language is a SOC system (Gromov and Migrinait 2017) and metaphor-motivated language is a sub-system of natural language use, it can be hypothesized that metaphorical language change is a SOC state that generates metaphorical expressions.

## 3 Phenotypes of metaphorical language change

With the hypothesis that metaphorical language change is a SOC, it can be inferred that the process should possess the SOC phenotypes, i.e. the features that are necessarily observed in a SOC state.

According to Watkins et al. (2016), the phenotypes of a SOC include *(i)* spatio-temporal fractal with power-law spatio-temporal correlations and *(ii)* self-organization, or self-tuning to the point of criticality. Phenotype *(i)* is observed in SOC systems because fractals are snapshots of systems operating at the SOC state (Bak and Chen 1989) and are characterized by self-similarity and power-law spatio-temporal correlations (Watkins et al. 2016). Phenotype *(ii)* means that a SOC system is not driven by an external force (Watkins et al. 2016), nor does it require a specification of initial conditions (Bak et al. 1988). Instead, it naturally, apparently and robustly evolves to the critical state, with a randomly initiated condition (Bak et al. 1989).

As metaphorical language change qualifies as a SOC, it should also possess the above two phtnotypes. It is expected to self-organize into a fractal with power-law spatio-temporal correlations and is characterized with self-similarity, scale-invariance, power distribution, and temporal inter-dependence. These features might be exhibited at the linguistic aspect as well as the aspect of conceptual mapping. Nevertheless, as the actuation problem is mainly concerned with linguistic expressions, the present study will verify the fractal nature of metaphorical constructions and the self-organization process observed in them, with an emphasis on the linguistic aspect rather than the aspect of conceptual mapping. It is possible to explore the fractal nature of conceptual mappings with metaphorical constructions because constructions are form-meaning pairs, but this is beyond the scope of the present study.

This section uses Chinese verb metaphors to illustrate and verify the above phenotypes. It first explains the data collection procedure, and then the fractal nature of the metaphorical constructions in terms of scale-invariance, power-law distribution and long-range temporal inter-dependence, and then the self-organization process of these constructions at the community level.

## 3.1 Data collection

The data collection procedure employs two measures to ensure the representativeness of the metaphorical constructions collected for the present study. One is random sampling. The other is thoroughness in data collection. Details are as follows:

– Randomly select 15 metaphor vehicle verbs from 105 modern Chinese verb metaphors compiled in Kang (2008). The random sampling is adopted so that the observations of the selected metaphors can be generalized to the category of Chinese verb metaphors;

– For each vehicle verb, collect all sentences containing the verb from a diachronic corpus that contains 59-year text data (1946-2004) of the Chinese newspaper *People's Daily* and then manually identify metaphorical instances among the sentences by referring to metaphorical senses defined in *CiHai*, which excludes three metaphor vehicles from further analysis because two of them are already conventionalized in 1946 and one lacks data[1];

– Use Stanford Parser (in Stanford CoreNLP 2.9.2) to parse those metaphor instances into dependency trees;

– For the list of dependency trees of each vehicle term, use DepCluster (Tang 2017) to obtain collostructions (Gries and Stefanowitsch 2004; Gries 2019; Stefanowitsch and Gries. 2003) and then manually group them to obtain representative constructions[2] of the verb metaphor.

The data of the twelve metaphors and their representative construction are summarized in Table 1. Both the construction number and the instance number in the data will be used for analyses in Section 3.3. We argue that the analyses based on the data of these metaphors yield proximal description of how verb metaphors emerge in Modern Chinese because these metaphors are randomly sampled from a large set of verb metaphors, their occurrence data are thoroughly collected from the large-scale diachronic corpus, and the constructions included in the analyses are representative of their usage in the language. For instance, it can be observed that verbs of different frequency are included in the present study. In Table 1, the frequency ranges from 86 to 7200, including both low and high frequency.

[1] The corpus is segmented with ICTCLAS (version 2011). *Cihai* (version 6) is a prestigious dictionary published by Shanghai Lexicographic Publishing House in 2009.

[2] DepCluster obtains collostructions by clustering dependency trees, fully explained in Tang (2017). The study also argues that the collostructions generated by DepCluster give a holistic profile of a lexeme's usage. All data are available for download from the author's personal web page.

**Table 1 Statistical summary of the 12 Chinese verb metaphors**

| | | | | |
|---|---|---|---|---|
| **Vehicle Verb** | *chao'fry* | *chonglang'surf* | *chongci'sprint* | *shache'brake* |
| **Constr. Number** | 10 | 7 | 9 | 7 |
| **Instance Number** | 769/880 | 63/96 | 174/193 | 135/154 |
| **Vehicle Verb** | *qiake'jam* | *zhuangche'collide* | *tuise'fade* | *tiaochao'(job)-hop* |
| **Constr. Number** | 8 | 6 | 8 | 11 |
| **Instance Number** | 123/149 | 72/86 | 344/359 | 195/210 |
| **Vehicle Verb** | *miaozhun'aim-at* | *jingen'follow* | *liangxiang'pose* | *lianyin'marry* |
| **Constr. Number** | 9 | 8 | 10 | 9 |
| **Instance Number** | 2297/2408 | 7124/7200 | 1467/1587 | 615/726 |

Notes: Constr. stands for construction. The first number in Instance Number denotes the instances included in typical constructions, while the second is the whole collected from the corpus. Some are excluded by DepCluster as outliers and atypical use.

## 3.2 Metaphor as a spatio-temporal fractal

The metaphorical-language-change-is-a-SOC hypothesis predicts that metaphorical language change generally takes the shape of a fractal. In perceiving language as a complex, adaptive and dynamic system, Larsen-Freeman and Cameron (2008: 109-111) argue that language use is fractal, citing Zipf's law as a piece of evidence because the law reveals the existence of a self-similar pattern between occurrences of the most frequent words at different text scales. Several studies (e.g. Altmann 1980; Hrebíček 1994; Hrebíček 1998; Andres 2012) in quantitative linguistics have also explored the fractal character in the mutual relationship between constructs and their constituents in languages. In line with these studies, the metaphorical expressions brought forth by a metaphor also form a spatio-temporal fractal. Spatially, these expressions should possess the following two characteristics: *(i)* the semantic function of a metaphor remains invariant despite the scale variance of its constructions, and *(ii)* the constructions of a metaphor follow a power distribution; temporally, the emergence of these metaphorical constructions is regulated by long range temporal interdependence manifested in the form of transformational rules.

### 3.2.1 Scale-invariant functional similarity

By its definition, fractals are self-similar, i.e. the parts of a fractal are similar to the fractal as a whole (Hrebíček 1994), despite that the fractal is larger in scale then the parts composing it. In natural language, this feature of scale-invariance can be observed in terms of linguistic expressions and contexts in which these expressions are used (Shanon 1993). In metaphorical language change, this feature is observed in the scale-invariant functional similarity of metaphorical expressions. That is, the semantic function performed by all metaphorical expressions of one metaphor remains similar (sometimes even invariant), despite that the scales (measured by constituent number) of these expressions change so as to adapt to different contexts.

Take for instance the metaphor COLLABORATION IS MARRIAGE denoted by the Chinese verb *lianyin'marry* in (5)[3], which means *collaborate with* in the sentence. Following the procedure in Section 3.1, nine constructions are obtained from 37 collostructions and 615 instances of the metaphor (Table 2)[4]. As argued before, these are regarded typical constructions of the metaphor and are representative of the metaphor's usage in modern Chinese, each construction covering more than four percent of the instance data.

| | | | | | | | | |
|---|---|---|---|---|---|---|---|---|
| (5) | *gaidi* | *baoxian* | *gongsi* | *zhudong* | *yu* | *qixiang* | *bumen* | *"**lianyin**".* |
| | local | insurance | company | proactively | with | climate | department | marry |

'Local insurance companies proactively "**marry**" climate departments.'

---

[3] All Chinese examples are taken from the diachronic corpus mentioned in Section 3.1.

[4] 726 instances are obtained from the corpus. After clustering with DepCluster, 111 instances are treated as outliers because they do not belong to any cluster that has more than 6 instances and are not included for study. The slot names in Table 2 are dependency types. Please refer to Appendix I for explanations.

**Table 2 Representative metaphorical constructions of *lianyin'marry***

<table>
<tr><th>No.</th><th>Construction</th><th>C.N.</th><th>F.O.Y</th></tr>
<tr><td>1</td><td>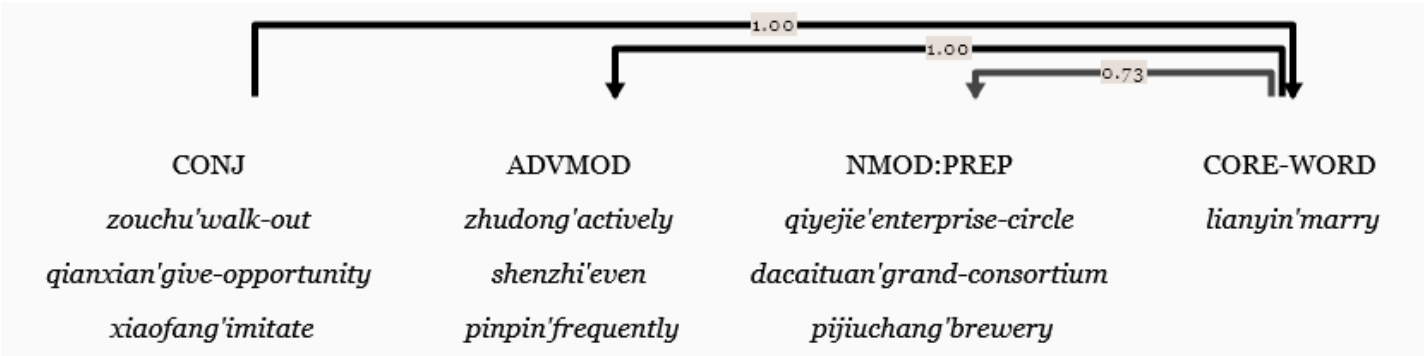
</td><td>3</td><td>1981</td></tr>
<tr><td>2</td><td>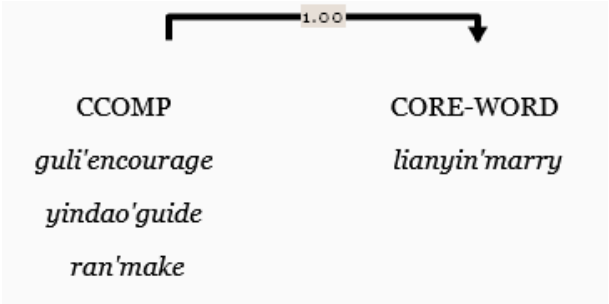
</td><td>1</td><td>1982</td></tr>
<tr><td>3</td><td>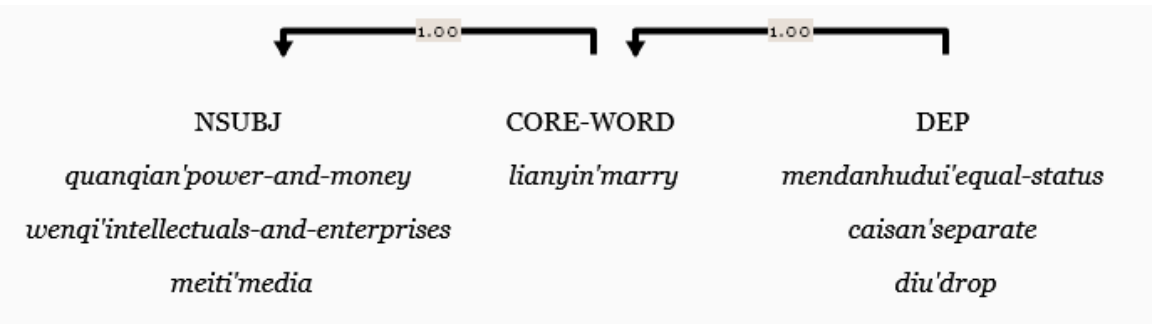
</td><td>2</td><td>1984</td></tr>
<tr><td>4</td><td>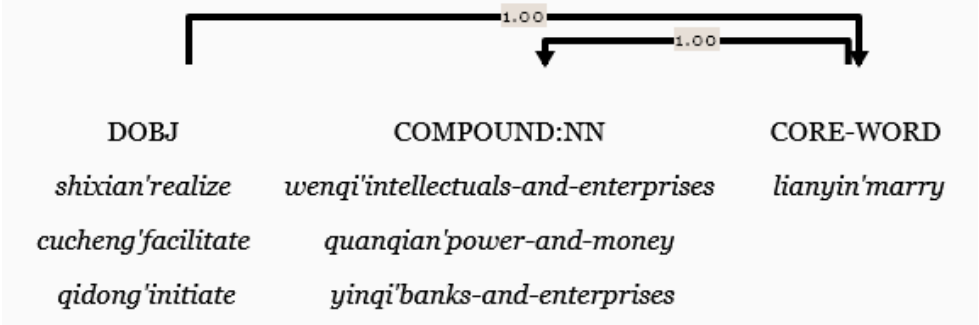
</td><td>2</td><td>1985</td></tr>
<tr><td>5</td><td>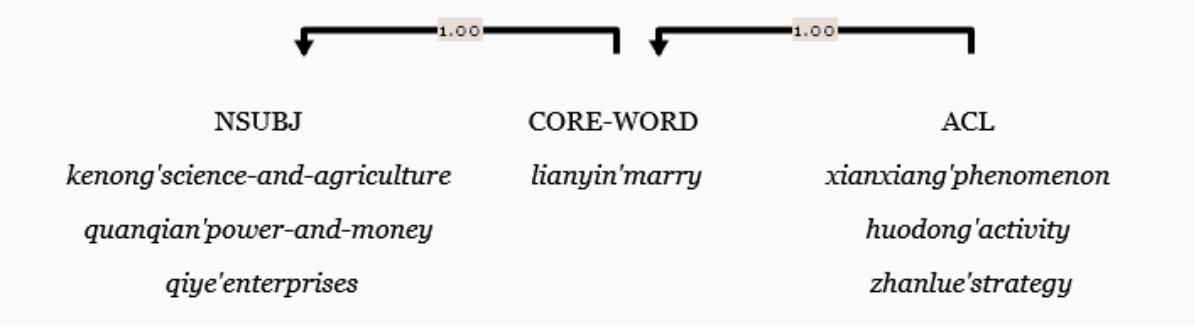
</td><td>2</td><td>1986</td></tr>
<tr><td>6</td><td>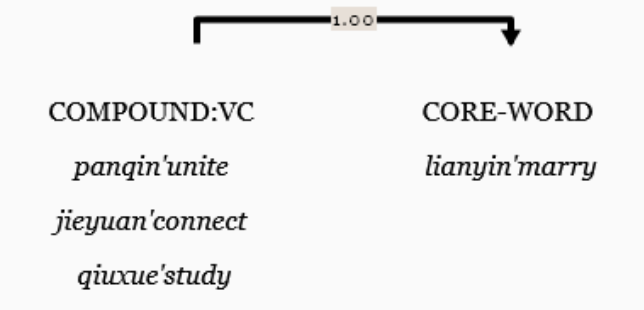
</td><td>2</td><td>1986</td></tr>
<tr><td>7</td><td>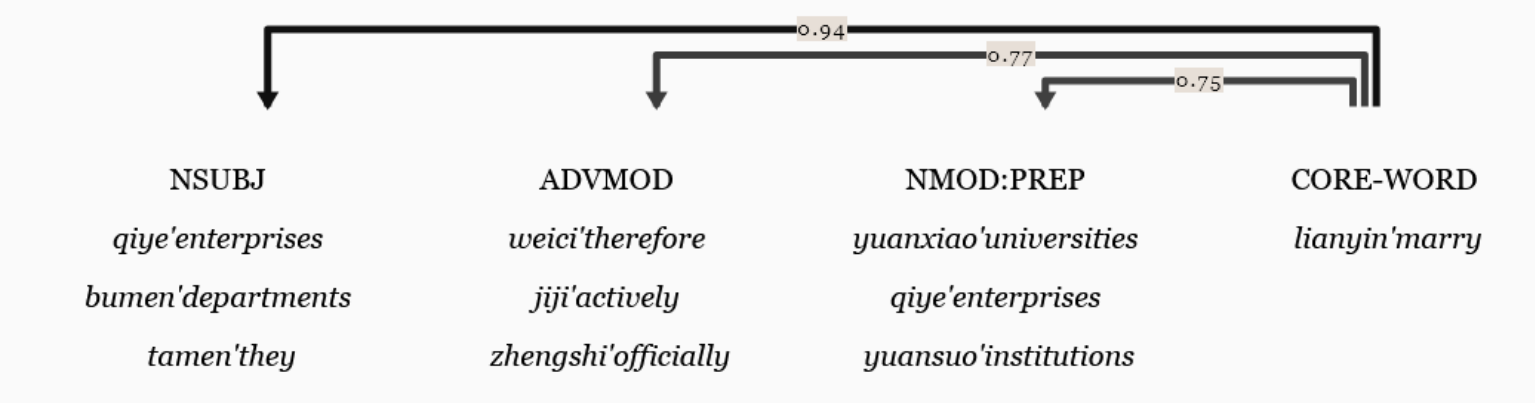
</td><td>4</td><td>1988</td></tr>
<tr><td>8</td><td>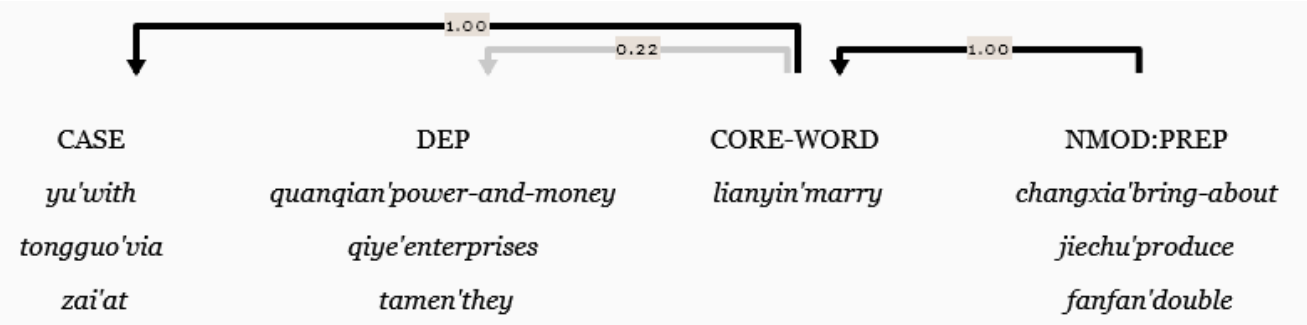
</td><td>3</td><td>1988</td></tr>
<tr><td>9</td><td>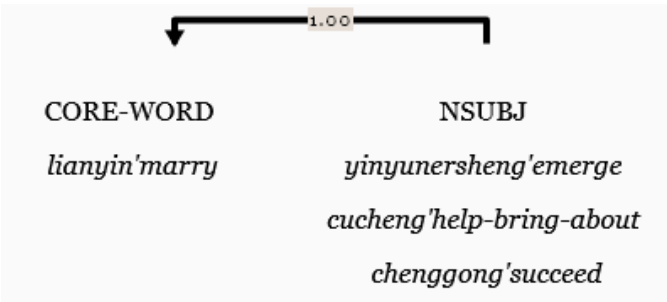
</td><td>1</td><td>1989</td></tr>
</table>

Notes: C.N. stands for constituent number, and F.O.Y. for First Occurrence Year.

The scale-invariant functional similarity of the metaphor is exemplified by the constructions in the table. These constructions surely are similar in semantic function because they are of the same metaphor, but the number of constituents that form each construction varies according to the grammatical function of the vehicle (denoted by CORE-WORD slot). They fall into two groups. The first group consists of No. 1 and No. 7, wherein the vehicle serves as the predicate verb of a conjunctive clause (No. 1) or the predicate verb of a main clause (No. 7). Both constructions take adjuncts such as adverbial modifiers and prepositional modifiers, which facilitate the understanding of the vehicle by evoking metaphorical mappings. Consequently, their constituent numbers are relatively bigger. No. 1 has three constituents and No. 7 has four.

In the second group, the vehicle no longer functions as a predicate verb in the main clause. Instead, it serves as a constituent inside a modifier (the head of attributive clause in No. 5), or a constituent inside a prepositional phrase (No. 8), or a constituent inside a verb phrase (such as the head of a complement clause in No.2), or the head of a topic (No. 3)[5], or the head of an objective clause (No. 4), or the head of a subjective clause (No. 9). In No. 6, it is part of a compound phrase, which can be used as a subject or an object. In this group, adjunctive constituents that attach to the vehicle verb generally do not appear. The logical subject of the vehicle might occur (as in No. 3, No. 4, No. 5 and No. 8) or not (as in No. 2, No. 6 and No. 9). As a result, the numbers of constituents in these constructions are smaller than the first group, varying between two and one.

The same phenomenon is observable in the data of the other eleven metaphors in Table 1. Similar to *lianyin'marry* metaphor, each of these metaphors has several constructions with various numbers of constituents to express the same metaphorical meaning. Scale-invariant self-similarity

[5] Topic sentence is a special syntactic pattern in Chinese. It has a topic besides the subject. Please refer to (6) for illustration.

is a common phenomenon in metaphorical language change motivated by Chinese verbs.

### 3.2.2 Power-law distribution

Earlier studies have obtained the insight that fractals and power-law distributions are tightly connected. For instance, Brown et al. (2002) argue that emergent fractals can be mathematically characterized by the power function below[6]:

$$y = ax^{-b} \qquad (1)$$

wherein $\boldsymbol{y}$ is a dependent variable, $\boldsymbol{x}$ is the independent variable, $\boldsymbol{a}$ and $\boldsymbol{b}$ are constants. Studies on power-law distributions also reach the consensus that power-law distributions are essentially the signature of fractals, because power-law distributions are also self-similar and scale-invariant and no other distributions possess such properties (Feldman 2012: 215-18; Larsen-Freeman and Cameron 2008: 109). In language studies, the connections between power-law distribution and fractals are also observed in Ninio (2011) who studies syntactic capacity in children's development, and in Evans (2020) who reports a long-tail, asymptotic curve of power-law distribution of clausal density.

Therefore, one way to verify that the constructions of a metaphor is a fractal is to show that they follow a power-law distribution. Using the power function in Equation (1), the present study applies non-linear regression to the frequency of the constructions of the twelve metaphors. The result is listed in Figure 2. Again, take *lianyin'marry* for illustration. The frequency of all constructions of the vehicle is obtained and sorted in descending order. Non-linear regression with

[6] The formula is adapted from Ward and Greenwood (2007).

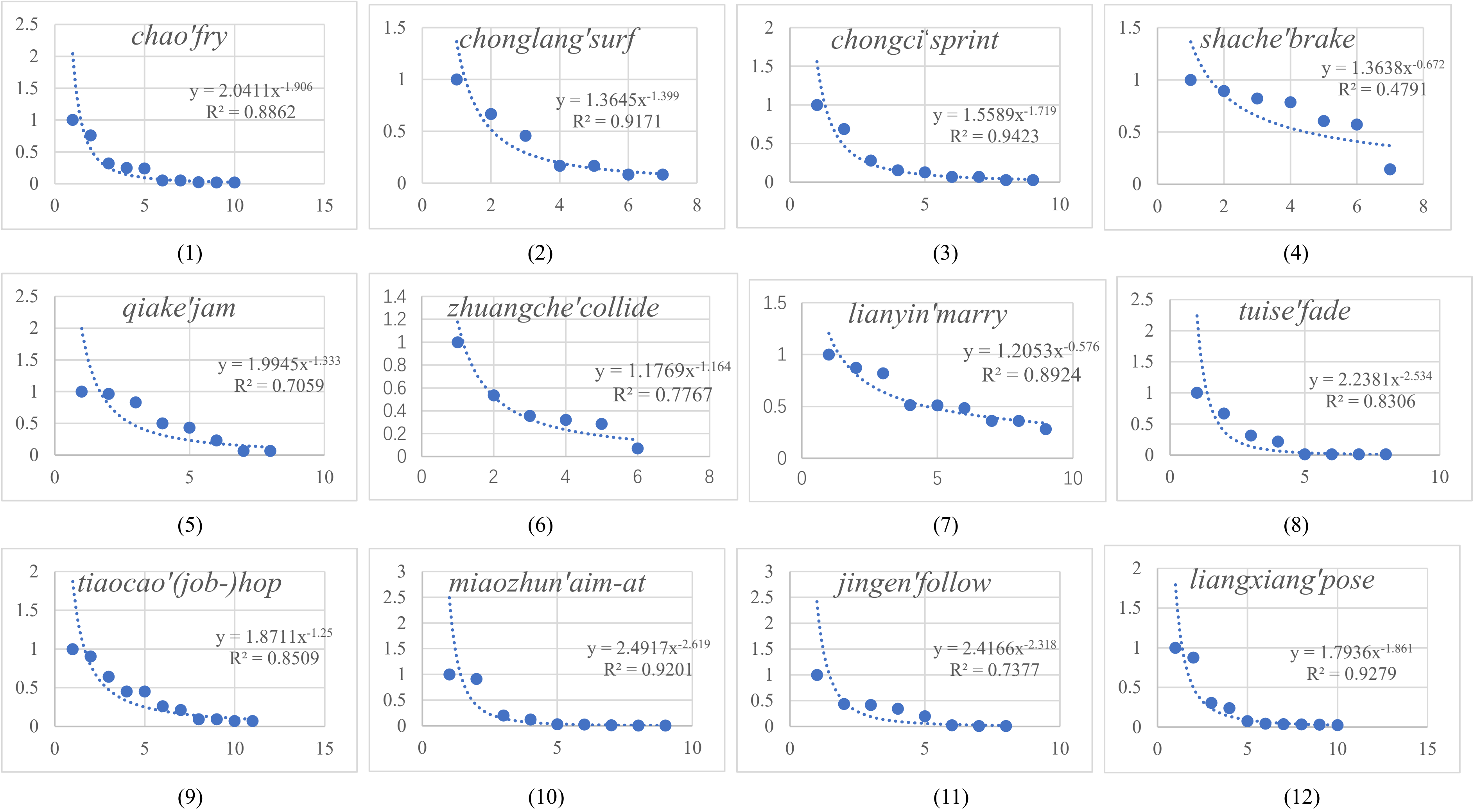

**Figure 2 Non-linear regression of the twelve metaphor constructions with power-law function**

Equation (1) is then applied to obtain the result given in sub-figure No. 7 in Figure 2. The determination coefficient of the regression ($\mathbf{R^2}$ in the sub-figure) is 0.8924, indicating that about 90% of the data obey power-law distribution. The average determination coefficient of the 12 metaphors is 0.8222, with the maximum 0.9423 and the minimum 0.4791. Therefore, it can be concluded that the constructions of each of the twelve metaphors are of power distribution. As power-laws are the signature of fractals, it can be inferred that metaphorical constructions of these verb metaphors are fractals. Furthermore, because these verbs are randomly selected, such inference can be generalized to the category of Chinese verb metaphors.

That metaphorical constructions exhibit a power-law distribution can also be explained as a manifestation of Menzerath-Altmann Law (MA Law) at work. That is, the constructions of a metaphor have a power distribution because they have to satisfy the constraints imposed by MA Law in a language. One piece of evidence for this observation is the identicalness of the mathematical formation of MA Law to Equation (1). The complete ML Law is defined in Equation (2) below (Altmann 1980; Eroglu 2014):

$$y(x) = Ax^{b}e^{-cx} \qquad (2)$$

wherein $\boldsymbol{A}$, $\boldsymbol{b}$ and $\boldsymbol{c}$ are model parameters. Because the exponential factor $\boldsymbol{e^{-cx}}$ seems to be irrelevant (Köhler et al. 2008: 282), a truncated formula is proposed, given in Equation (3) below (Andres 2010; Andres et al. 2012; Benešová and Čech 2015; Köhler et al. 2008: 282):

$$y(x) = Ax^{b},\ b < 0 \qquad (3)$$

This equation is identical with Equation (1) above.

The identicalness of Equation (3) to Equation (1) allows an interpretation that the constructions of a metaphor follow a power distribution because they need to comply with MA Law. Consider the juxtaposition of (5) above and (6) below. Either (5) or (6) can be considered as one semantic construct. According to the MA Law, the longer the construct, the shorter its constituents (Eroglu 2014). Example (5) is a shorter semantic construct as it has only one clause. Nevertheless, it uses eight words to express the *lianyin'marry* metaphor. Example (6) is a longer semantic construct because it has four clauses. Among them, the verb clause *lianyin'marry* (should be an *ING*-form if translated into English) expresses the same metaphor as that of (5), but it consists of only the vehicle word itself and its understanding relies on speakers' language experience. The clause contains only one word because it occurs inside a longer semantic construct with four clauses. To recapitulate, the metaphorical expression of *lianyin'marry* in (6) is more concise than that in (5) because it has to comply with the MA Law.

(6) *"**lianyin**" jiajie, shuo-qilai rongyi, zuo-qilai keyao dongzhengge-de*

**marry** graft talk-ASP easy do-ASP require serious-action-ASP

'In regard to **marrying** and grafting, it is easier said than done and requires serious actions.'

Hrebíček (1994) discusses how the relation between MA Law and fractals relates to syntactic hierarchy, given below:

> … the Menzerath-Altmann law holds for all possible language constructs and their constituents; to be a language construct means, therefore, to have constituents in the sense of the law, and thus constructs represent two different levels (or subsystems); the Menzerath-Altmann law is derivable from the expression valid for self-similar fractals. (Hrebíček 1994: 86)

Similar opinions are also found in Hrebíček (1998) and Andres (2010). Thus, the metaphorical constructions complying with MA Law should fall into different syntactic hierarchical levels. In regard to Chinese verb metaphors exemplified by (5) and (6), these two levels relate to verb nominalization. Verb clauses of *lianyin'marry* as in (5) are far from nominalization. But when they are nominalized, they behave much like nouns as in (6). This will be further discussed in the next section.

### 3.2.3 Long-range temporal interdependence

The assumption that metaphorical language change is a fractal also predicts that there is long-range temporal interdependence among metaphorical constructions. This prediction provides a direct answer to the actuation problem of metaphorical language change. Explicit specification of long-range temporal interdependence will help explain why some metaphorical expressions should occur earlier than some others, as is demonstrated below in the case of Chinese verb metaphors.

A theoretical explanation of long-range interdependence among metaphorical expressions is found in the correlation between power distribution and $1/f$ scaling in cognition. In cognitive studies, power distributions are interpreted as $1/f$ scaling (Kello et al. 2011; Van Orden et al. 2003; Van Orden et al. 2005), which characterizes the criticality state of cognition where the probability density of all sizes of events is a power function (Ward and Greenwood 2007), the same function as Equation (1). $1/f$ scaling occurs in two possible situations (Van Orden et al. 2003): *(i)* a single process extending to multiple timescales; *(ii)* multiple processes linked across multiple timescales. In both cases, such fractal-like scaling suggests that there are simple rules of organization that govern the emergence of the system (Brown et al. 2002).

Metaphorical language change, which is motived by the cognitive process of metaphorical mapping, should fall into situation *(i)* and be treated as one single cognitive process extending to multiple timescales. In the process, a single metaphor is repeatedly employed in a speech community at multiple time scales and with different forms in its course of conventionalization. Like other $1/f$ scaling cognitive phenomena, the emergence of observable constructions of the metaphor is also governed by simple rules.

These simple rules should encompass the whole process of metaphorical language change. Studies in SOC show that the rules of long-range temporal interdependence do not simply describe periodical behaviors, but reflect the whole process (Kello et al. 2010; Van Orden et al. 2003; Van Orden et al. 2005). The variation of the whole is present in each of its parts (Van Orden et al. 2003). For anything that changes over time, its next state is determined by its previous state and the rules that govern how states change over time (Van Orden and Stephen 2012). Each rule is the function of the system's entire history (Jensen 1998: 9). Therefore, to understand the rules constraining metaphorical language change, the linguistic changes triggered by one metaphor should be treated as one process, as is argued by Gibbs and Van Orden:

> A critical state entails all the propensities to speak that satisfy constraints due to relevant history and present context. Relevant historical constraints include the relevant anatomical and psychological constraints … the details of the speakers' previous history, … as well as what has been said previously in the exchange. (Gibbs and Van Orden 2012: 14)

The above observation is verified in the case of the twelve Chinese verb metaphors. Explicit rules can be computed from their construction data, using the time information associated with each of the constructions (as is illustrated in Table 2). The procedure to obtain these rules are explained below:

– For each typical construction of each of the twelve metaphors, record the year of the earliest instance of the construction as first-occurrence-year;

– For the list of constructions of each metaphor, if the first-occurrence-year of $\boldsymbol{C_1}$ is smaller than that of $\boldsymbol{C_2}$, increase the frequency of the potential transformation rule $\boldsymbol{C_1} \rightarrow \boldsymbol{C_2}$ by one;

– Use the above two steps to examine the constructions of all the twelve metaphors and obtain frequency $\boldsymbol{F}$ for all $\boldsymbol{C_i} \rightarrow \boldsymbol{C_j}$, and then compute the conditional probability $P(C_j|C_i) = \frac{P(C_i \rightarrow C_j)}{P(C_i)}$.

Using the filters $Frequency(C_i \rightarrow C_j) > 5$ and $P(C_j|C_i) \geq 0.8$, we obtain five transformation rules, listed in Figure 3. Each rule consists of two constructions, and each construction is denoted by a dependency and is illustrated with an example sentence below. CCOMP(X, VEHICLE) is exemplified by (7), NSUBJ(X, VEHICLE) by (8), CONJ(X, VEHICLE) by (9), DEP(X, VEHICLE) by (6), and ROOT(ROOT, VEHICLE) by (5).

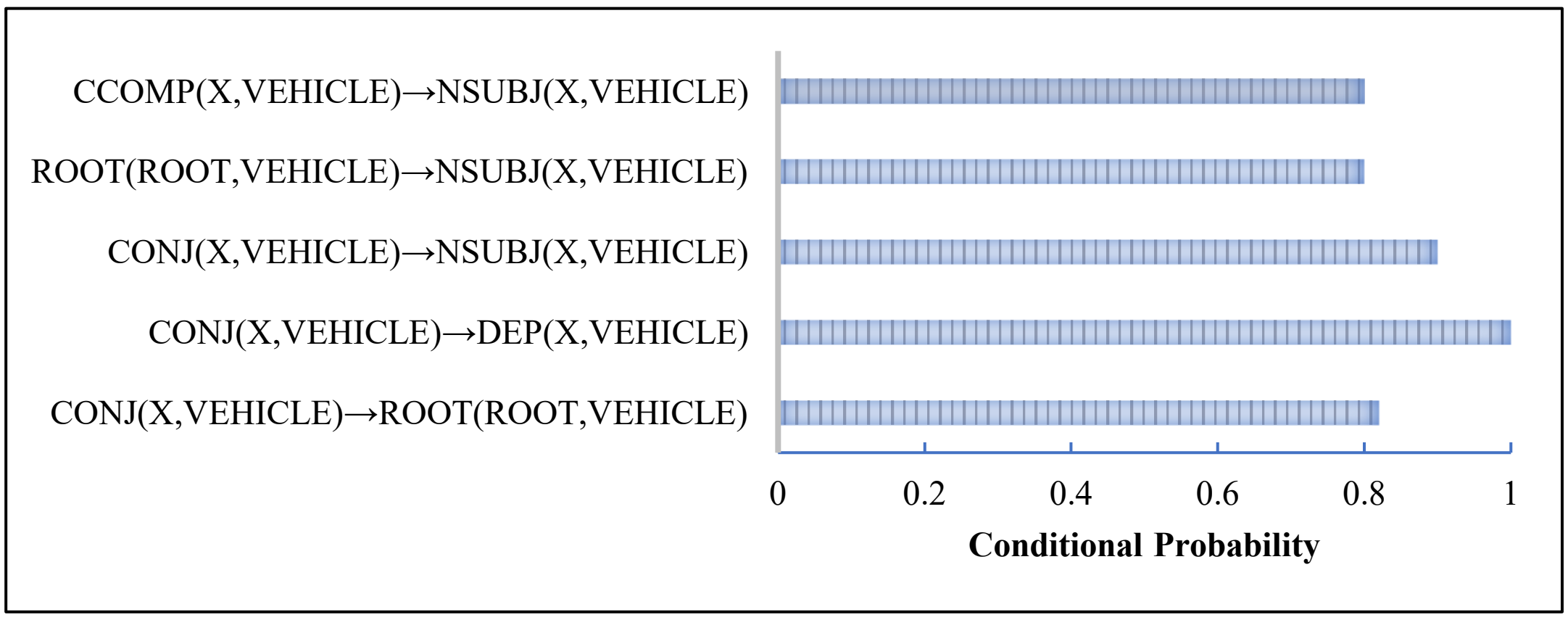

**Figure 3 Transformation rules and their conditional probability**

(7) *tamen* *xiwang* *yu* *kejirenyuan* ***lianyin***. (CCOMP)

they wish with technicians marry

'They wish to **marry** technicians.'

(8) ***lianyin*** *yingyun'ersheng-le.* (NSUBJ)

marry emerge-ASP

'**Marriage** emerged.'

(9) *tamen qianxin* *zhuanyan jishu* *zhudong* *yu* *yuanxiao* ***lianyin***. (CONJ)

they devotedly study technology actively with universities marry

'They are devoted to technology studies and actively **marry** universities.'

The conditional probability of rule "CONJ(X, VEHICLE) → DEP(X, VEHICLE)" is 1.0, indicating that of all constructions of the twelve metaphors, CONJ(X, VEHICLE) precedes DEP(X, VEHICLE). Using (9) and (6) for illustration, the rule stipulates that if a metaphor is to be used in topicalization (see [6]), the vehicle has to be used as predicate verb in conjunction with another verb (see [9]). A more general hypothesis is that Chinese speakers might possess such knowledge of transforming CONJ(X, VEHICLE) to DEP(X, VEHICLE). Similar interpretation also applies to the rule "CONJ(X, VEHICLE) →ROOT(ROOT,VEHICLE)", in which ROOT(ROOT,VEHICLE) denotes that the vehicle is used as predicate verb in a sentence, as is exemplified by (5).

The other three rules in Figure 3 all involve the construction NSUBJ(X, VEHICLE) and have high conditional probability ($P(C_{nsubj}|C_i)$≥0.8), listed below:

CCOMP(X, VEHICLE) → NSUBJ(X, VEHICLE)

ROOT(X, VEHICLE) → NSUBJ(X, VEHICLE).

CONJ(X, VEHICLE) → NSUBJ(X, VEHICLE)

As illustrated in (8), the vehicle in NSUBJ(X, VEHICLE) is the head of a subjective clause. These three rules essentially are rules of nominalization in Chinese, stating that the nominalization of a verb metaphor in modern Chinese requires knowledge of the vehicle being used inside a complement clause, or as a predicate verb alone, or as the predicate verb in conjunction with another verb.

The transformation rules obtained above are similar to the concept of transformation in Chomsky (1957), whose tenet is that every adult possesses relatively few simple sentence patterns and a complex set of transformation rules that combine and modify simple sentences into the infinite number of complicated sentences (Hayes 1967). Nevertheless, the functions of the rules obtained in the present study are much narrower and more explicit — they are to perform nominalization, transforming verb phrases into nominal phrases. In building complicated sentences, these rules possess two important properties. The first is self-similarity. Each of the transformation rules changes the order of its components, omitting some of them and (sometimes) adding grammatical markers, but the semantic structure of the newly produced construction remains similar to the original. The second is contraction. These transformation rules tend to drop some complements and use only those essential components, forming more contract constructions, as is discussed in Section 3.2.1.

These transformation rules are explicit manifestation of long-range temporal interdependence in language change motivated by Chinese verb metaphors. With these rules, it is possible to predict that some metaphorical expressions should occur before or after some other expressions, thus providing partial resolution to the actuation problem of metaphorical language change.

## 3.3 Metaphor as a self-organization process

The second phenotype is that a metaphor self-organizes to the point of criticality, without an external force nor an initial condition. The self-organization process of individual language speakers in using a metaphor is already proposed and explained in the discourse dynamics approach to metaphor and the dynamic view of metaphor (Gibbs and Colston 2012; Gibbs 2013). Gibbs and Santa Cruz (2012) propose an explicit model for the process, focusing on the role of frequency and the shift to higher order (or second order, explained in the next paragraph). This model is illustrated in Figure 4 using the conceptual metaphor SOCIAL ORGANIZATIONS ARE PLANTS, (1), and (4). The interpretation of (1) (*they had to prune the workforce*, left side of the figure) establishes multiple links among neural, cognitive, linguistic, cultural, and evolutionary levels. More use of the metaphor leads to the strengthening of the links and the generation of attractors, i.e. regions a system tends to settle (Lass 1997: 293) or phase spaces frequently occupied by a system (Gibbs and Santa Cruz 2012). In the figure, the attractors are denoted by heavier links at the right side.

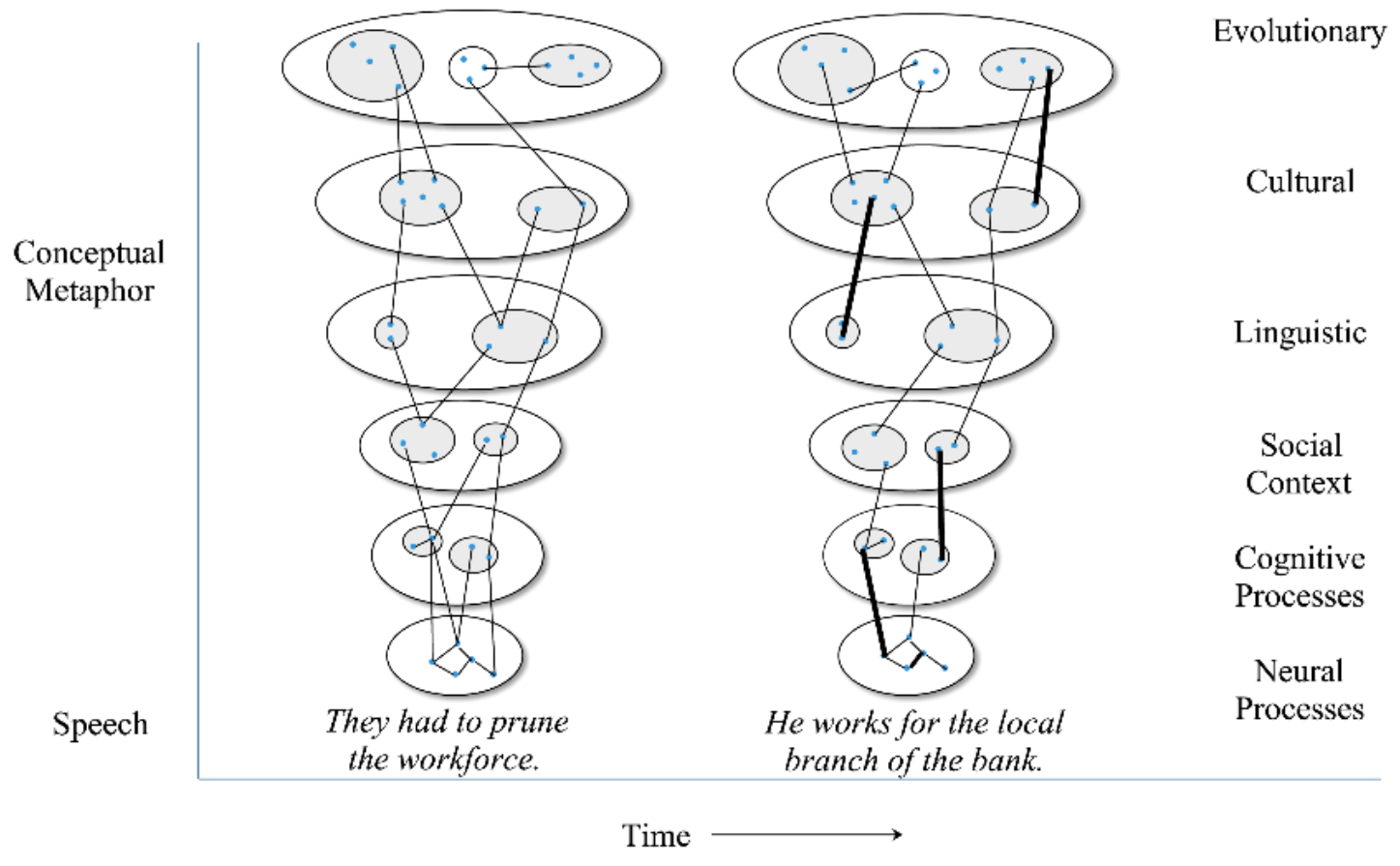

**Figure 4 Self-organizing process of conceptual metaphor interpretation**[7]

[7] The figure is adapted from Gibbs and Santa Cruz (2012).

The model also depicts a shift from the first-order linguistic activity to the second-order linguistic activity. First-order linguistic activities are real-time, contextually determined behavior with semiotic significance (Love 1990; Love 2004), constrained by first-order contextual constraints (Juarrero 1999). For instance, the use of (1) in Figure 4 can increase the likelihood of another instance of the metaphor in similar contexts, which in turn increases the likelihood of other instances of the metaphor. This kind of impact of one instance upon another is called first-order contextual constraint. Second-order linguistic activities are abstraction of first-order linguistic activities in terms of words, sentences, and meanings etc. and these abstract descriptions are generalizations arising from first-order activities (Love 1990; van den Herik 2017). These analytic descriptions emerge as contextual constraints on the components, which can be called second-order contextual constraints (Juarrero 1999). In Figure 4, the attractors that emerge as contextual constraints on their components are second-order contextual constraints. They are more like regularities and are used in the top-down fashion in the understanding process. Example (4) (*he works for the local branch of the bank*) is assumed to be understood with second order contextual constraints. The time arrow at the bottom of the figure denotes the shift from first-order constraints to second-order constraints: in the early phase of development, a metaphorical expression is more likely understood via first-order constraints; in later phases, second-order constraints emerge from frequent occurrence of instances and they are more likely to be used to understand metaphorical expressions.

The statistics obtained in the present study yields empirical evidence for the role of frequency and the shift to second order in the self-organization process. The role of frequency is observed in the statistical analysis of the constant $\boldsymbol{b}$ in Equation (1). Via constant $\boldsymbol{b}$, a correlation between the scope of perturbation and frequency can be established. On the one hand, the constant $\boldsymbol{b}$ relates to the manner of perturbation such as the scope of perturbation (Jensen 1998: 39-42). In the sandpile model[8], if extra sand grains are added at random positions of the surface of the pile, the system should yield $1/\boldsymbol{f}^2$ spectrum, with $\boldsymbol{b} = 2$. However, if the extra grains are added only along the closed walls, the system should yield $1/\boldsymbol{f}$, with $\boldsymbol{b} = 1$. This is further illustrated in Figure 5, which plots the curves generated by Equation (1) with ***a*** = 2 and ***b*** changes from .5 to 2.5. It can be observed that as ***b*** increases, the curve becomes steeper, meaning that the perturbation scope becomes narrower and the distribution is more dominated by some types. On the other hand, the data given in Table (3-4) relate ***b*** to the frequency of the twelve metaphors. Table 3 provides related information of the twelve metaphors, including the ***b*** value, frequency of the metaphors in the corpus, and first-occurrence-year. Table 4 gives the Pearson correlation coefficient between ***b*** value and frequency computed from Table 3. The correlation coefficient is 0.52968, showing that ***b*** value is positively related to frequency.

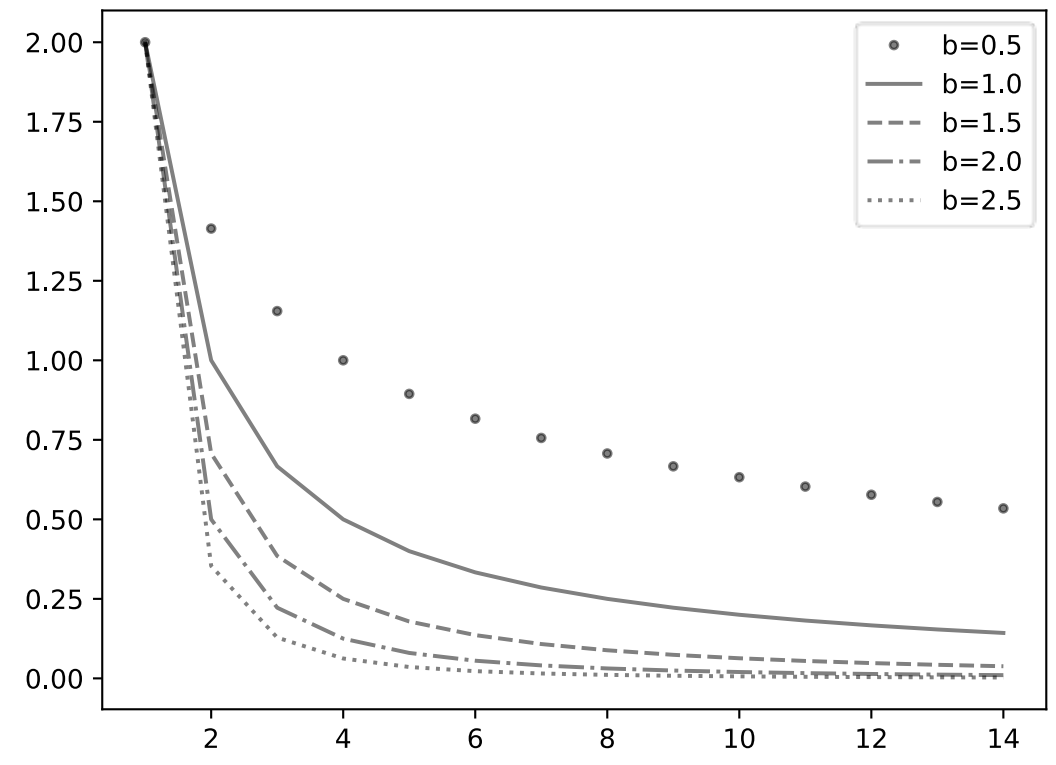

**Figure 5 The relation between *b* and perturbation in power-law function with *a* = 2**

[8] Sandpile model is used to explain SOC in Bak, Tang, and Wieserfeld (1988) and Bak, Chao, and Wiesenfeld (1987).

**Table 3 Information of *b*, first-occurrence-year (FOY), and frequency of 12 metaphors**

| No. | Vehicle Term | ***b*** | FOY | Frequency |
|---|---|---|---|---|
| 1 | *chao'fry* | 1.906 | 1964 | 880 |
| 2 | *chonglang'surf* | 1.399 | 1987 | 96 |
| 3 | *chongci'sprint* | 1.719 | 1957 | 193 |
| 4 | *shache'brake* | 0.672 | 1955 | 154 |
| 5 | *qiake'jam* | 1.333 | 1964 | 149 |
| 6 | *zhuangche'collide* | 1.164 | 1963 | 86 |
| 7 | *lianyin'marry* | 1.136 | 1981 | 726 |
| 8 | *tuise'fade* | 2.534 | 1949 | 359 |
| 9 | *tiaocao'hop* | 1.25 | 1988 | 210 |
| 10 | *miaozhun'aim-at* | 2.619 | 1952 | 2408 |
| 11 | *jingen'follow* | 2.318 | 1947 | 7200 |
| 12 | *liangxiang'pose* | 1.861 | 1960 | 1587 |

**Table 4 Pearson Correlation Coefficient with *b*, first-occurrence-year (FOY), and frequency**

| | ***b*** | FOY | Frequency |
|---|---|---|---|
| ***b*** | 1 | | |
| FOY | -0.554363698 | 1 | |
| Frequency | 0.529680999 | -0.459786631 | 1 |

Because scope of perturbation is positively related to ***b*** and ***b*** is positively related to frequency, the scope of perturbation is positively related to frequency. This inference can partly explain the variation of the number of constructions in Table 1. Take for instance the data of No. 1 and No. 5 in the table. No. 1 is the metaphor introduced by vehicle term *chao'fry*. It has higher frequency and higher ***b*** value than No. 5. Higher frequency implies broader perturbation scope, i.e. a wider range of constructions should be involved in the metaphorical language change. Accordingly, No. 1 has 10 metaphorical constructions while No. 5 has 8, as are given in the table. In other words, more frequent use of a metaphor in the self-organization process should lead to more complex linguistic patterns of the metaphor.

The shift from first-order constraints to second order constraints can be observed in the negative correlation (-0.554) between first-occurrence-year (FOY) and $b$ in Table 4. Theoretically, the correlation between first-occurrence-year and $\boldsymbol{b}$ can be established by conventionalization. A smaller first-occurrence-year indicates a longer time of conventionalization in the corpus, leading to higher degree of conventionality and more likelihood of shifting from first-order constraints to second-order constraints. As second-order constraints carry normative force and can shape or inform first-order linguistic activities (van den Herik 2017), speakers' behavior shall be much more alike to each other. Consequently, metaphorical expressions used by the speakers should become more dominant. Some metaphorical constructions tend to dominate the distribution, leading to a higher $\boldsymbol{b}$ value (see Figure 5). On the contrary, a larger first-occurrence-year indicates a more recent emergence, lesser time of conventionalization, lower degree of conventionality and less likelihood of shift from first-order constraints to second-order constraints. Accordingly, the distribution of the constructions should also be less dominant and the $\boldsymbol{b}$ value is smaller. To recapitulate, the shift from first-order constraints to second-order constraints leads to higher degree of conventionality of a metaphor, which in turn leads to more dominant distributions of its expressions, or a higher $\boldsymbol{b}$ value.

# 4 Conclusions

In advocating the view of language as complex systems, Larsen-Freeman and Cameron (2008: 109) poses the following question: "what is the shape of a language trajectory in the landscape or state space of human language-using potential?" Metaphorical language change is a category of language trajectory in the landscape of human language-using potential. The present study proposes that a resolution to the actuation problem of metaphorical language change is to profile this type of language trajectory as a SOC state. Literature in the view of language as complex systems and the dynamic view of metaphor shows that metaphorical language change is motivated by non-linear interaction among language speakers and is characterized by intermittent and avalanche-like changes. These features qualify metaphorical language change as a SOC state. Accordingly, the shape of metaphorical language change can be described with the phenotypes of SOC, given below:

– Metaphorical expressions introduced by a metaphor self-organize into a self-similar and scale-invariant fractal that possesses a power-law distribution and exhibits long-range cognitive interdependence manifested as transformation rules;
– Attractors (metaphorical constructions) are formed in the self-organizing process of metaphorical language change, driven by frequency;
– There is a shift from first-order constraints to second-order constraints in the process;

These features are illustrated and verified in the paper with statistical analyses of twelve randomly selected Chinese verb metaphors in the diachronic corpus.

A scientific theory based on a finite number of observations often seeks to relate the observations and to predict new phenomena by constructing general laws (Chomsky 1957: 49). The metaphorical-language-change-is-a-SOC hypothesis also seeks to predict the developmental trajectory of novel Chinese verb metaphors in the present study. Those obtained transformation rules are speculated to be valid in predicting the behavior of Chinese verb metaphors because cares are taken to ensure the representativeness of the data used for analyses. In addition, it is speculated that the metaphorical-language-change-is-a-SOC hypothesis also applies to other types of metaphors and metaphors in other languages because the hypothesis is based on the theoretical basis of the view of language as complex systems and the dynamic view of metaphor. It might be observed that language change motivated by metaphors in general form self-similar scale-invariant fractals, and that simple transformation rules can also be obtained for these metaphors, although empirical evidence is yet to be obtained.

The insights on metaphorical language change obtained in the present study contribute to the view of language as complex systems and the claim that language is fractal. Although scholars have alluded to the fractal nature of language for some time, the discussion has been limited, relying almost exclusively on the fractal as a metaphor in their description and explanation (Evans 2020). The extent to which fractal patterns appear in language use is still uncertain (Larsen-Freeman and Cameron 2008: 111). The present successful application of fractal analyses to metaphorical language change supports the claim in Evans (2020) that fractal analysis is a fruitful tool for discerning the dynamic relationships among the multiple components of complex systems as they interact over time. SOC might as well be applicable in exploring other kinds of language change, such as metonymic language change.

The insights concerning power distributions and transformation rules of metaphorical change are useful in studies of metaphor acquisition and computational metaphor studies. For instance, the transformation rules in Figure 3 are not only rules of verb metaphor nominalization, but also the acquisition order of metaphorical constructions. Such knowledge stipulates that the use of nominalized verb metaphors in novel metaphor should not be encouraged because this will add to cognitive burden in metaphor acquisition. In the field of automatic metaphor recognition, the knowledge of transformation rules will help narrow the scope of metaphor recognition. If there are general rules transforming novel metaphorical constructions to more conventionalized metaphorical constructions, it might be sufficient to limit metaphor recognition to the scope of novel metaphors in building knowledge database of conceptual mapping. The transformation rules can then be used to obtain those more conventionalized metaphorical expressions.

## Appendix I Dependency types

| Name | Explanation |
|---|---|
| ACL(H, D) | D is the head of an attributive clause of H. |
| ADVCL:LOC(H, D) | D is the head of an adverbial clause of H. |
| ADVMOD(H, D) | D is an adverbial modifier of H. |
| CASE(H, D) | D is a grammatical marker of H. |
| COMP(H, D) | D is the head of a complement of H |
| COMPOUND:XX(H, D) | D forms a compound phrase with H. XX indicates the syntactic category of the compound. |
| CONJ(H, D) | H is a conjunction of D and H occurs before D in word order. |
| DEP(H, D) | D forms a dependency relation with H, but the nature of dependency is not clearly specified. |
| DOBJ(H, D) | D is the direct object of H. |
| MOD:PREP(H, D) | D is the head of a clause/phrase that modifies H. The clause or phrase is introduced by a preposition. |
| NMOD:PREP(H, D) | D is a prepositional modifier of H. |
| NSUBJ(H, D) | D is the nominal subject of H. |
| ROOT(ROOT, D) | D is the predicate verb of the sentence. |